\documentclass[runningheads]{llncs}
\usepackage{graphicx}
\usepackage{bm}
\usepackage{amsmath,amssymb}
\usepackage{multirow}
\begin{document}
\title{Feature Selection and Hyperparameter Fine-tuning in Artificial Neural Networks for Wood Quality Classification\thanks{The authors are grateful to FAPESP grants \#2016/06538-0, \#2018/02822-1 and \#2019/07825-1.}}

\author{Mateus Roder\inst{1}\orcidID{0000-0002-3112-5290} \and
Leandro Aparecido Passos\inst{1}\orcidID{0000-0003-3529-3109} \and
Jo\~{a}o Paulo Papa\inst{1}\orcidID{0000-0002-6494-7514} \and \\
Andr\'e Luis Debiaso Rossi\inst{2}\orcidID{0000-0001-6388-7479}}
\authorrunning{Roder Author et al.}
\titlerunning{FS and Hyperparameter Fine-tuning in ANNs for Wood Classification}
\institute{Department of Computing, S\~ao Paulo State University\\
	Av. Eng. Luiz Edmundo Carrijo Coube, 14-01, Bauru, 17033-360, Brazil \\
\email{\{mateus.roder, leandro.passos, joao.papa\}@unesp.br}\and
Department of Production Engineering, Paulo State University\\ 
	Rua Geraldo Alckmin, 519 - Vila Nossa Sra. de Fatima, Itapeva, 18409-010, Brazil
\email{andre.rossi@unesp.br}}
\maketitle              
\begin{abstract}
Quality classification of wood boards is an essential task in the sawmill industry, which is still usually performed by human operators in small to median companies in developing countries. Machine learning algorithms have been successfully employed to investigate the problem, offering a more affordable alternative compared to other solutions. However, such approaches usually present some drawbacks regarding the proper selection of their hyperparameters. Moreover, the models are susceptible to the features extracted from wood board images, which influence the induction of the model and, consequently, its generalization power. Therefore, in this paper, we investigate the problem of simultaneously tuning the hyperparameters of an artificial neural network (ANN) as well as selecting a subset of characteristics that better describes the wood board quality. Experiments were conducted over a private dataset composed of images obtained from a sawmill industry and described using different feature descriptors.  The predictive performance of the model was compared against five baseline methods as well as a random search, performing either ANN hyperparameter tuning and feature selection. Experimental results suggest that hyperparameters should be adjusted according to the feature set, or the features should be selected considering the hyperparameter values. In summary, the best predictive performance, i.e., a balanced accuracy of $0.80$, was achieved in two distinct scenarios: (i) performing only feature selection, and (ii) performing both tasks concomitantly. Thus, we suggest that at least one of the two approaches should be considered in the context of industrial applications.
\keywords{Artificial Neural Network Optimization  \and Wood Quality Classification \and Sawmill Problem \and Hyperparameter Tuning \and Feature Selection.}
\end{abstract}
\section{Introduction}
\label{s.introduction}

Industries have experienced many technological advances in recent years, resulting in more complex processes, systems, and products. As a consequence, the management of integrated manufacturing processes and operation analyses are crucial to delivering high-quality products to clients. In this scenario, the quality of raw materials is also paramount for high quality manufactured products. However, imperfection detection, as well as quality classification of raw materials, such as woods in sawmill companies, usually is still performed by trained human operators~\cite{abdullah2007}. Notwithstanding, the process is inherently subjective, since it is a visual analysis, and these experts may suffer from fatigue after a long working period performing repetitive activities. Consequently, it is expected the increased number of incorrect classifications~\cite{vieira2016}. 

\begin{figure*}[!htb]
  \centerline{
    \begin{tabular}{ccc}
	\includegraphics[width=4cm, height=1.5cm]{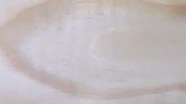} &
	\includegraphics[width=4cm, height=1.5cm]{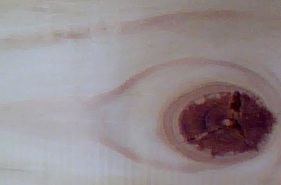}  &
	\includegraphics[width=4cm, height=1.5cm]{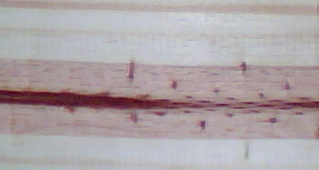} \\
	Wood quality A. & Wood quality B. & Wood quality C.
    \end{tabular}}	
    \caption{Three different qualities of wood boards (A, B, and C) according to company's rule.}
  \label{fig.wood_quality}
\end{figure*}

These disadvantages stimulated the scientific community towards the implementation of visual inspection systems, aiming to perform defect and quality classification autonomously. Therefore, this work focuses exclusively on Machine Learning (ML) techniques that have been successfully employed for these tasks on wood boards~\cite{gu2009}. Recently, some studies investigated the performance of different ML techniques to classify the quality of wood surface~\cite{roder2017,qi2018}. These studies have used data generated from images captured in a real sawmill company and classified by a specialist in three levels of quality, according to the company rules, i.e., zero defect is found in the wood piece (A), only small defects, such as knots, are found (B), and defects that compromise the quality of the product, such as groups of knots and exposed pith (C). Fig.~\ref{fig.wood_quality} depicts some examples of wood images classified at each level.

Despite the success obtained in these studies, each ML algorithm has its inherent tendency towards data specificity, which influences the model's induction and, thus, its predictive performance. Therefore, one can adjust such tendencies through a proper hyperparameter (HP) selection. The task of finding the best HP values is known as \textit{hyperparameter tuning} and usually aims at improving the model's predictive performance while keeping the model as simple as possible. Although some HP values may fit sufficiently well different kinds of problems, it is a common practice to search for the hyperparameter that provides the best solutions concerning each problem at hand \cite{roder2017,qi2018}.

Besides, another challenge in the context of this work is extracting the images' more representative features, i.e., the features that best describes the problem, since the more descriptive they are, the higher the effectiveness of the technique. Regarding classification tasks, features are usually extracted through image descriptors, such as statistical measures from the Gray Level Co-occurrence Matrix (GLCM) \cite{haralick1973} and Local Binary Patterns (LBP)~\cite{ojala1996}. However, many of these features may be correlated to each other or do not add any relevant information for the ML technique. Therefore, the process of selecting a subset of these features, referred to as feature selection (FS), can be applied to select the most descriptive ones. Moreover, since features are specific for each problem, FS has to be carried out for each data set separately.

Considering that both the HP tuning and the task of feature selection relies upon each other, they should be performed simultaneously to generate more robust models, concerning generalization purposes. In general, ML techniques require tuning more than one HP, since the tackled problems generally are described by many features, thus implying on large search spaces. In this context, metaheuristics approaches are commonly employed to solve such problems by randomly initializing a collection of candidate solutions, which interact among themselves and perform a directed exploration of the search space toward the results that best fit a desirable target function with an acceptable computational cost. Such approaches are commonly employed to solve problems related to ML techniques hyperparameter tuning~\cite{roder2021reinforcing,de2021fine,passosSIBGRAPI:2017} and feature selection~\cite{pereiraECCOMAS:19}, among others~\cite{cao2021new,rodrigues2020adaptive}.

In this paper, we investigate the problem of FS and Artificial Neural Network (ANN) hyperparameter tuning applied in the context of wood boards quality classification.  Experiments were carried out using the population-based metaheuristic Particle Swarm Optimization (PSO)~\cite{kennedy1997,roder2020harnessing} to simultaneously perform both tasks over a Multilayer Perceptron (MLP) ANN. Moreover, the results compared against five distinct baselines, as well as a random search, confirms the relevance of the proposed approach. We hypothesize that the predictive performance of ANN models can be improved since they depend on the HP values and the set of features used to describe the problem.

Therefore, the main contributions of this paper are twofold: (i) to propose a method capable of simultaneously selecting the hyperparameters that best performs over an MLP network as well as selecting the subset of features that best describe each image sample, and (ii) to foster the scientific community regarding material and wood quality classification. The remainder of this paper is presented as follows. Section ~\ref{s.related} defines the problem of hyperparameter tuning and feature selection, and provides a brief description of some related works. Section~\ref{s.theoretical} presents the main concepts of ANN and PSO. The experimental methodology employed to evaluate the effects of FS and MLP hyperparameter tuning over the models' performance is described in Section~\ref{s.methodology}. Results are presented and discussed in Section~\ref{s.results}, and finally conclusions are presented in Section~\ref{s.conclusions}.

\section{Problem Definition and Related Work}
\label{s.related}

Hyperparameter tuning and feature subset selection are two widely employed tasks carried out in the data mining context, aiming to improve models' predictive performance as well as simplifying them. Therefore, this section formalizes the problem of simultaneously performing these two tasks. Besides, it presents an overview of studies related to wood quality classification and the importance of  HP tuning and FS.

\subsection{Problem Definition}

The problem investigated in this paper consists of tuning the HP values of an MLP Artificial Neural Network algorithm, as well as selecting a subset of features that are relevant for the problem of wood quality classification towards the improvement of the models' predictive performance.  

\begin{sloppypar}
Let $A$ be an MLP algorithm that comprises the hyperparameter space $\Lambda$. For each hyperparameter setting $\lambda \in  \Lambda$, let $A_{\lambda}$ represent the learning algorithm $A$ that employs the hyperparameter setting $\lambda$. Also, consider $D = \{(x_1,y_1), (x_2,y_2), \ldots, (x_n,y_n)\}$ a dataset composed of $n$ instances, such that $\bm{x} \in \mathbb{R}^{m}$ is a feature vector and $y$ is the target value. Moreover, one can define $\kappa$ as the subset of feature from $\bm{x}$. Finally, let $A^{\kappa}$ be an algorithm $A$ trained with a subset of features $\kappa$~\cite{luo2016}. 
\end{sloppypar}

\begin{sloppypar}
Therefore, the main goal of HP tuning is to finding  $\lambda^* = \text{arg min}_{\lambda \in \Lambda} M(A_\lambda, D) \in \Lambda$ that minimizes some loss function, such as the misclassification rate using the algorithm $A$ over instances not used for training purposes. Moreover, one can estimate the misclassification rate $M(A,D)$  achieved by $A$ when trained and tested on $D$ through a stratified multi-fold cross-validation resampling method. 
\end{sloppypar}


Similarly to HP tuning, the goal of feature subset selection is to find $\kappa^*=\text{arg min}_{\kappa\subseteq\bm{x}}M(A^\kappa, D) \subseteq \bm{x}$ which minimizes the loss function achieved by $A$ when trained on $D$.


Therefore, the aim of combining hyperparameter tuning and feature subset selection is to find the hyperparameter setting  $\lambda^* \in \Lambda$  and the feature subset $\kappa^* \subseteq \bm{x}$ which has the lowest misclassification rate among all HP settings and features subsets, i.e., $(\kappa^*,\lambda^*) =  \text{arg min}_{\lambda \in \Lambda, \kappa \subseteq \bm{x}} M(A^{\kappa}_{\lambda}, D)$.


\subsection{Related Work}

One of the first studies to investigate the problem of wood quality control using an automated visual inspection system based on machine learning techniques was accomplished by~\cite{pham1996}. Since then, many others have investigated this problem aiming to improve predictive performance by analyzing and selecting different features, as well as optimizing HP values of ML techniques~\cite{tiryaki2014,roder2017}. 


Tiryaki et al.~\cite{tiryaki2014} employed an ANN for modeling the wood surface roughness in the machining process. The study highlights some variables that have impact and influence in the surface roughness, such as wood species, the feed rate, the number of the cutter, and the cutting depth. The model's predictive performance was good enough to allow its application in the wood industry in order to optimize effort, time, and energy.

Others addressed the problem of combining FS and HP tuning for ML techniques over different applications. As previously mentioned, some techniques are more sensitive to HP tuning and FS than others. Besides, some optimization methods, such as metaheuristic approaches based on evolutionary algorithms and swarm intelligence~\cite{kennedy2001}, for instance, have successfully accomplished the task. 



In this context, Roder et al.~\cite{roder2017} used the PSO algorithm to tune the HP of an ANN applied for wood quality classification aiming to enhance the model's predictive performance. The authors employed the GLCM~\cite{haralick1973} to extract features from the same dataset used by~\cite{affonso2017}, which is composed of five statistical measures for two angles, i.e., $0$ and $90$ degrees: entropy, energy, maximum intensity, inverse difference moment, and correlation. Experimental results obtained up to $6\%$ of accuracy gain concerning the ANN classification, and corroborate the necessity of tuning the ANN hyperparameters; stating the efficiency of PSO for such a task. 

\section{Theoretical Background} 
\label{s.theoretical}

This section briefly introduces the main concepts of the techniques employed in this work, i.e., the Multilayer Perceptron neural network and the Particle Swarm Optimization algorithm, as well as the process of wood image feature extraction.


\subsection{MultiLayer Perceptron ANN}
\label{subsec:ann}

\begin{figure}
  \centering
  \includegraphics[scale=0.28]{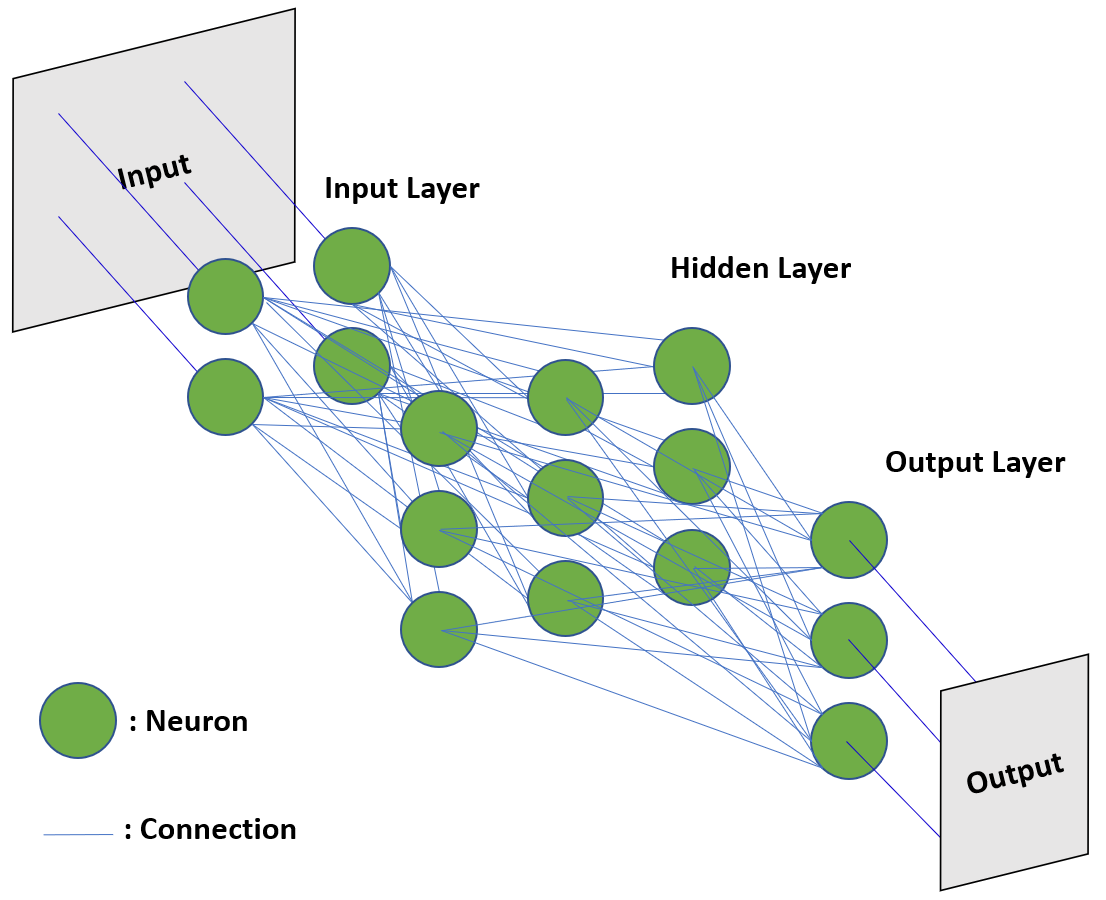}
  \caption{MultiLayer Perceptron representation.}
  \label{fig:mlp-3d}
\end{figure}

MLPs are composed of an input and an output layer, as well as one or more hidden layers of neurons, which can be fully or partially connected. A neural network is called fully connected when each neuron from a given layer is connected to all neurons of the next one. Similarly, it is said to be partially connected when some neurons of adjacent layers are not connected.  These connections are represented by a weight matrix, which is usually adjusted through gradient-based learning algorithms. With a single hidden layer, an MLP is capable of representing a large number of functions, thus sufficient for the purpose of this study. Figure~\ref{fig:mlp-3d} depicts the model architecture.

The conventional algorithm to train an MLP network is the backpropagation, which is composed of the forward pass, which exposes the input data to a series of linear operations followed by non-linear activations, and the backward phases, which is responsible for propagating the output error and update the network weights.

\subsection{Particle Swarm Optimization} 
\label{sec:pso}


Particle Swarm Optimization~\cite{kennedy1995} is a global optimization technique based on the social behavior of birds, fishes, and insects, among others. The method comprises a swarm composed of a set of individuals capable of sharing information among themselves concerning their positions in the search space, as well as the relative quality of this position, denoted by a fitness function.


\begin{sloppypar}
In short, each PSO particle is represented by its current position, velocity, and the best position found during the training process. The position of a particle $i$ is represented by a point in a $D$-dimensional space, given by $\Psi_{i} = \{\psi_{i1},\psi_{i2}, \ldots,\psi_{iD}\}$. Further, the particle  velocity is defined by $\bm{v}_i=\{v_{i1},v_{i2},\ldots, v_{iD}\}$, and finally the best position found by this particle is represented by $\bm{p}_i=\{p_{i1},p_{i2},\ldots,p_{iD}\}$. Besides, the best position found among all particles is represented by $\bm{p}_g$.
\end{sloppypar}

A particle will move in a particular direction depending on its current position, velocity, and best position. Additionally, it also depends on the best position found by the other particles in the swarm. Therefore, the position of a particle $\psi_{ij}(t+1) = \psi_{ij}(t) + v_{ij}(t)$ is computed for each dimension $j\in\{1,2,\ldots,D\}$ at time step $t$. Further, the velocity $v_{ij}(t)$ is updated using the following equation:


\begin{equation}
v_{ij}(t+1) = \Gamma \cdot v_{ij}(t) + \varphi_1 \cdot r_1 \cdot (p_{ij} - x_{ij}(t)) +  \varphi_2 \cdot r_2 \cdot(p_{gd} - x_{ij}(t)),
\label{equ:partVelocidade}
\end{equation}
where $\Gamma$ denotes the inertial weight, introduced by~\cite{shi1998} to balance the global and local search, $r_1$ and $r_2$ are two independent values uniformly distributed in the range $[0, 1]$, and $\varphi_1$ and $\varphi_2$ are acceleration constants.

Such a representation of a particle is suitable for hyperparameter fine-tuning, which considers real- and integer-valued numbers. However, it is not adequate for feature selection tasks since it requires a categorical or binary representation. Therefore, this work employs a variation of the method, adapted for feature selection, as discussed in Section~\ref{s.methodology}.

\subsection{Feature Extractors}
\label{sub:backg_feature}



Haralick et al.~\cite{haralick1973} proposed a set of mathematical tools to extract statistical features from images using a Gray Level Co-occurrence Matrix. The GLCM is capable of describing the frequency of occurrences in grayscale transitions for an image in a pixel-by-pixel fashion. Further, the features are extracted considering the relationship of each pixel with its neighbors over four different angles, i.e., $0$, $45$, $90$, and $135$ degrees, where $0$ and $90$ degrees are the most employed ones. Moreover, the model is capable of extracting a total of $14$ statistical features, named Angular Second Moment, Contrast, Correlation, Variance, Inverse Difference Moment, Sum Average, Sum Variance, Sum Entropy, Entropy, Difference Entropy, Information Measures of Correlation ($1$ and $2$) and the Maximal Correlation Coefficient.

Another well-known texture descriptor is the Local Binary Pattern~\cite{ojala1996}, which converts the image to a gray-scale level and performs a pixel-by-pixel comparison over the entire image considering a selected number of neighbors. In this comparison, the central pixel of the square formed by its neighbor assumes the value $1$ if it is greater or equal to its neighbors, or $0$ otherwise. Such value is stored in an array, which is further employed for converting the binary intensity code to a decimal value for the pixel at hand.

Afterward, the same computation is performed for all pixels compounding the image. Further, it is generated a histogram of the distribution of the values, which will compose the final vector describing the image. Notice the method was initially proposed with the number of neighborhood fixed in $3\times 3$. Later, some changes allowed LBP to deal with a larger number of neighborhoods, resulting in a non-square structure, usually employing a circular pattern since it only requires the definition of the radius, instead of the $N x N$ arrangement. 







\section{Methodology}
\label{s.methodology}

This section presents the methodology concerning the material and methods employed during the experiments. It briefly describes the datasets, the process of feature extraction, the modeling of the hyperparameter fine-tuning and feature selection processes using PSO, the methods used for evaluation purpose, and the baselines considered for comparison.

\subsection{Dataset}
\label{subsec:data}

This work employs a dataset $D$ composed of features extracted from $374$ instances of wood board images obtained in a Brazilian sawmill~\cite{vieira2016}. As stated in the problem definition, each instance $D=(\bm{x}_i, y_i)$ is composed of an $m$-dimensional feature vector $\bm{x}_i \in \mathbb{R}^{m}$, whose features were extracted using both GLCM and LBP, as described in Section~\ref{subsec:feature}, and a target value $y_i$, denoting the wood quality. Each sample's target value is established according to rules defined by the sawmill company, where ``A'' stands for a high-quality standard and comprises $144$ instances, ``B'' denotes an intermediate quality and comprises $177$ instances, and ``C'' represents lower quality, comprising $53$ samples. 


\subsection{Image Texture Descriptors}
\label{subsec:feature}

The feature set was obtained by joining the features extracted from two texture descriptors, namely, statistical measures extracted from GLCM and the LBP. While the statistical measures have the advantage of enabling the interpretation and comprehension of the image characteristics through different measures, LBP is robust in the treatment of gray-scale images, with a good performance for scale changes caused by illumination~\cite{ojala1996}.

Concerning the GLCM, this paper employed $0^o$ and $90^o$ to extract six measures: angular second momentum, energy, contrast, correlation, dissimilarity, and homogeneity, resulting in $12$ characteristics for each image. Besides, LPB uses $24$ neighbors as well as a radius of size $3$, resulting in $26$ characteristics for each image. Thus, joining GLCM and LBP features resulted in $38$ predictive attributes. 


\subsection{MLP Hyperparameter tuning and Feature Selection using PSO}
\label{subsec:hp_fs}

This work employed a fully connected network composed of a single hidden layer for the task of classification. Further, it also employed the PSO algorithm to fine-tuning the three principal hyperparameters of the model, namely the number of neurons in the hidden layer $\gamma$, the learning rate $\eta$, and the momentum term $\mu$.

As mentioned previously, the PSO algorithm was employed to perform a combined task of hyperparameter tuning and feature selection, hereafter referred to as HP-FS-PSO. Considering the former task, PSO decision variables are modeled admitting one integer values to represent the number of units in the hidden layer, as well as two real numbers to represent the learning rate and momentum.

Further, since the task of feature selection does not assume continuous representation, PSO requires some modifications to work properly in this context. The main change is made on the position representation, which must be treated as a result of probability analysis from particles' velocity to decide what features are relevant to the context~\cite{kennedy1997}. Therefore, considering the task of feature selection, each particle's decision variable is represented by a binary value, where $1$ means to consider a feature whereas $0$ means to discard a feature.




Since the particle's decision variables employed for feature selection assume binary values, it is necessary to binarize each position, $\psi_{ij}$ such that $\psi_{ij}=1$ if $s(v_{ij}) > r_{3}$ and $0$ otherwise. Notice $r_{3}$ is the threshold, a real number generated randomly in the range $[0,1]$, and $s(\cdot)$ is the logistic function.




Concerning the MLP hyperparameters' search configuration, the hidden layer size is optimized in the range $[2, 60]$, the learning rate and momemtum assume values in the range $[0, 1]$, and the feature subset selection is defined by a binary variable, i.e., assuming either $0$ or $1$. Additionally, it also presents the MLP hyperparameter default values used by Weka~\cite{hall2009}. The number of neurons in the hidden layer is defined as $\gamma=\frac{(\text{NA} + \text{NC})}{2}$, where $NA$ is the number of attributes and $NC$ is the number of classes. Thus, the default value for our problem, which has $38$ attributes and $3$ classes, is $20$ neurons.

%


Further, the PSO algorithm also has its own hyperparameters, defined as follows: Number of Particles $N=30$, Acceleration Constant 1 $\varphi_{1}=1.494$, Acceleration Constant 2 $\varphi_{2}=1.494$, Inertia Weight $\omega=0.729$, and Maximum Velocity $\upsilon$. Notice tuning such hyperparameters would lead to a ``never-ending'' problem. Therefore, these values were empirically selected based on similar works~\cite{kennedy1997,roder2017}.  Besides, the maximum velocity $\upsilon$ varies according to the upper limit of its respective hyperparameter.


The optimization process is performed until meeting the stop criterion, i.e., the maximum number of iterations, which was set in $300$.

\subsection{Evaluation}
\label{subsec:eval}

Metaheuristic algorithms guide their search for the best solutions according to the outcome of a fitness function. In this work, this measure is obtained from the predictive performance of the ANNs over a validation subset.



Moreover, since the dataset used in this work is composed of an imbalanced class distribution, i.e., there is a considerably smaller number of examples labeled as ``C'' class. Therefore, measures that acknowledge this imbalance are more suitable for the task. Thus, in this study, we considered the Balanced Accuracy (BAC)~\cite{brodersen2010}


The ANN training and evaluation were performed using a nested stratified k-fold cross-validation (CV) re-sampling method. Such an approach splits the dataset into $k=10$ partitions, where one of them is used to test, and the remaining folds are employed for training purposes. In this context, the training folds are used to train the model during the optimization process, i.e., finding the best MLP hyperparameters and the best subset of features. Therefore, PSO assesses the average BAC considering the fitness value over the validation set. Further, the best set of hyperparameters and features found in this process is them applied to train the model and induct the prediction of the testing set samples' labels. Such a process guarantees that the data used to evaluate the model is never used in the model training steps and, consequently, in the MLP hyperparameter tuning and feature selection processes.

Finally, due to the stochastic process of PSO, the optimization process was repeated during $10$ runs, aiming to perform a statistical analysis through the Wilcoxon signed-rank test~\cite{Wilcoxon:45} with $5\%$ of significance.

\subsection{Baselines}
\label{subsec:ref_app}

In order to evaluate and compare the results obtained by PSO, six baselines methods were compared in the context of the combined MLP hyperparameter tuning and feature selection:

\begin{itemize}
	\item Method 1 \textbf{(M1)} : Default hyperparameter values defined by Weka and the whole set of features;
	\item Method 2 \textbf{(M2)} : MLP hyperparameter tuned using PSO and the whole set of features;
	\item Method 3 \textbf{(M3)} : Default MLP hyperparameter values  defined by Weka and feature subset selected by PSO;
	\item Method 4 \textbf{(M4)} : Default MLP hyperparameter values  defined by Weka and dimensionality reduction performed by Principal Components Analysis (PCA);
	\item Method 5 \textbf{(M5)} : MLP hyperparameter values tuned by PSO and dimensionality reduction performed by PCA;
	\item Random Search \textbf{(RS)} : Random selection of MLP hyperparameter values and feature subset. This approach considered the same number of solutions evaluated by PSO.
\end{itemize}

The methodology adopted in this work aims at analyzing the PSO performance from different $x$ perspectives. First, M1 is the baseline for both tasks, i.e., the MLP hyperparameter using default parameters provided by Weka considering the whole set of features. Further, M2  allows analyzing PSO influence for the task of MLP hyperparameter tuning, with no feature selection, while M3 investigates the opposite, i.e.,  PSO influence to the task of feature selection with using MLP default hyperparameters. Moreover, M4 employs the default hyperparameter with a dimensionally reduced feature subset performed by PCA, while M5 combines MLP hyperparameters tuning using PSO with PCA. Finally, RS represents the analysis of random combinations of MLP hyperparameter values and selected features.

The experiments carried out in this work were coded using Python\footnote{https://www.python.org/} and R~\cite{r2014}. Further, the feature extraction task was implemented in Python using the Scikit-image\footnote{http://scikit-image.org/} package, while the MLP network was developed using the RWeka package in R, which is an interface to Weka. Finally, PSO was also implemented in R.
\section{Experimental Results}
\label{s.results}


This section presents the predictive performance of the ANNs assessed for the HP-FS-PSO method, as well as the baselines techniques. Notice the results are also provided with the $p$-values considering the Wilcoxon signed-rank test compared to the HP-FS-PSO method as the reference for statistical analysis purposes. Further, values presented in bold stand for the most accurate result overall.

\subsection{Optimization Evaluation}

Regarding the optimization performance over the validation set, one can observe in Table~\ref{tab:summary} that HP-FS-PSO and M3 have obtained the best results, achieving a  BAC average of $0.850$. Such techniques provided an improvement of around $10\%$ compared to M1, which represents an ANN using default HP values and the whole set of features. Therefore, the most important finding from these results is that the optimization of MLP hyperparameters, and even more a proper selection of the more suitable features, have a strong influence in the induction of MLP models applied to wood boards quality classification. On the other hand, results also show that performing only one task may also be enough to increase performance since M3 performed only the task of feature selection, while M2, which obtained an average BAC of $0.834$, performing only the task of MLP hyperparameter tuning.  Although by a narrow margin, in this case, the set of features was more expressive for the model predictive performance than tuning the network hyperparameters.

This behavior suggests that a user should, at least, select a set of features for the hyperparameter values defined \textit{a priori}. Besides the BAC values, Table~\ref{tab:summary} also provides the $p$-values considering the Wilcoxon signed-rank test compared to the HP-FS-PSO method as the reference. These $p$-values support our previous observations, considering only M3 and the random search obtained a significance level higher than $\alpha=0.05$.

\begin{table}[!ht]
\caption{Average BAC and the standard deviation concerning the task of MLP hyperparameter tuning and feature selection, considering the validation set over $10$ executions. Notice the ``$p$-values'' are compared against the HP-FS-PSO reference.}
\centering
\begin{tabular}{lcc}
\hline\noalign{\smallskip}
& Validation & \multirow{2}{*}{$p$-value} \\
& \scriptsize{(BAC$\pm$ SD)} & \\
\hline
M1 & 0.752 $\pm$ 0.008 &  0.002 \\
M2 & 0.834 $\pm$ 0.015 & 0.008\\
M3 & \textbf{0.850 $\pm$ 0.015} & 0.557 \\
M4 & 0.739 $\pm$ 0.011 & 0.002 \\
M5 & 0.825 $\pm$ 0.022 &  0.006 \\
RS & \textbf{0.845} $\pm$ 0.012 & 0.131 \\
HP-FS-PSO & \textbf{0.850} $\pm$ 0.011 & Ref.\\ 
\noalign{\smallskip}\hline
\end{tabular}
\label{tab:summary}
\end{table}

Notice the positive behavior of the random search approach, which is somehow expected since the model presents itself as more sensitive to a proper selection of the features instead of the network hyperparameter tuning, which is expected to be a more straightforward task due to the binary nature of the search space. Further, the data dimensionality reduction performed by PCA showed to be inadequate for this problem, as denoted by methods M4, which obtained BAC average results lower than using a default configuration, and M5. The main reason lies in the fact that PCA may not be able to describe sufficiently well the problem due to its linear nature.

\begin{figure}[!ht]
  \centerline{
	\begin{tabular}{cc}
		\includegraphics[width=6.1cm,height=5.5cm]{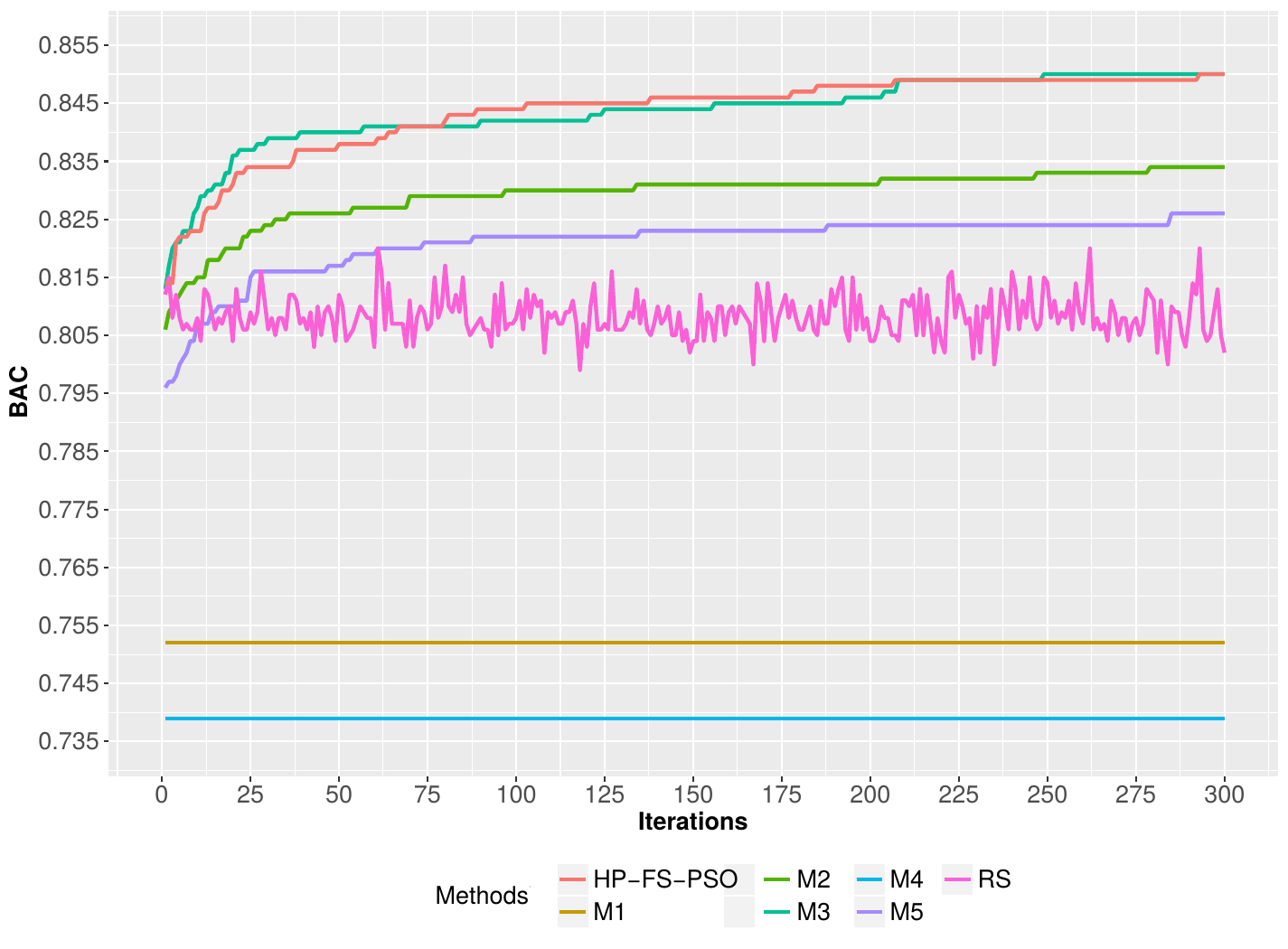}&
		\includegraphics[width=6.1cm,height=5.5cm]{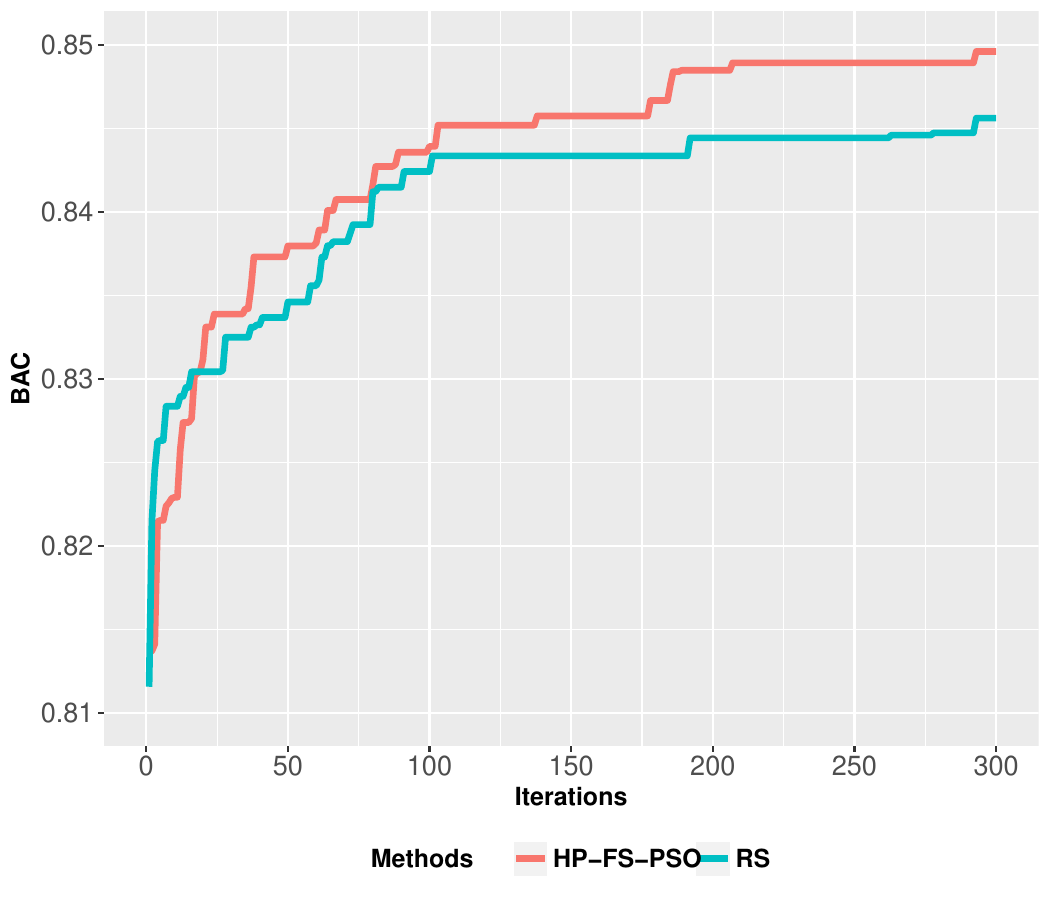}\\
		(a) & (b)
	\end{tabular}}
	\caption{PSO convergence considering the evaluation dataset (a) and evolution of the BAC values of HP-FS-PSO and RS. In this case, the best BAC value found by RS up to a iteration is kept in the next iterations (b).}
	\label{fig:all_methods} 		
\end{figure}

For a  better understanding of each method's behavior during PSO convergence, Figure~\ref{fig:all_methods} depicts the optimization performance (a) and the evolution of the BAC values (b) during $300$ iterations. Figure~\ref{fig:all_methods} (a) considers the average values over $10$ runs, where each iteration in the RS curve reflects the average evaluation among $30$ executions, i.e., the same number of assessments performed by PSO considering $30$ particles. Finally, M1 and M4 are represented by fixed lines since no optimization was performed over such approaches. Figure~\ref{fig:all_methods}(b) corroborates our claim that HP-FS-PSO performed a guided search through the MLP hyperparameters and features spaces, improving its performance over the iterations. Notice PSO can reduce the number of iterations required for finding reasonable BAC values since it obtained relatively high accuracies (around $0.840$) after $80$ iterations only.

Besides, Figure~\ref{fig:all_methods}(b) depicts the HP-FS-PSO performance compared against a random search considering the best results over each iteration, instead of an average.  One can observe the random search performed slightly better during the first $30$ iterations. Afterward, the HP-FS-PSO surpassed RS and kept this advantage until reaching the $300$ iterations. As previously mentioned, it is possible to note that there is a considerable improvement of BAC for both methods in the first $100$ iterations, and then there is a slowdown in the BAC growth. Such a piece of information is of extreme relevance for industrial applications, since it may save time and effort during the task of tuning the model's hyperparameter selecting the best subset of features.

\subsection{Classification}

This section investigates the ANN generalization power by evaluating the predictive performance of the model over the testing set, considering the best set of hyperparameters and a subset of features found during the optimization process. Table~\ref{tab:summary_test} presents the BAC values obtained in this context. Notice the values presented in bold stand for the most accurate approach overall.

\begin{table}[!ht]
\caption{Average BAC and the standard deviation concerning the task of MLP wood quality classification considered the testing samples and the best set of hyperparameters and subfeatures found during the optimization process for each approach. Notice the ``$p$-values'' are compared against the HP-FS-PSO reference.}
\centering
\begin{tabular}{lcc}
\hline\noalign{\smallskip}
& Test & \multirow{2}{*}{p-value} \\
& \scriptsize{(BAC$\pm$ SD)} & \\
M1 & 0.755 $\pm$ 0.077 & 0.275\\
M2 & 0.779 $\pm$ 0.042 & 0.547\\
M3 & \textbf{0.798} $\pm$ 0.087 & 0.722\\
M4 & 0.705 $\pm$ 0.057 & 0.010\\
M5 & 0.773 $\pm$ 0.071 & 0.232\\
RS & 0.745 $\pm$ 0.080 & 0.131\\
HP-FS-PSO & 0.794 $\pm$ 0.062 & Ref.\\ 
\hline
\end{tabular}
\label{tab:summary_test}
\end{table}

In general, the results of test data are in agreement with those of the validation set. The baseline method M3 led to the best BAC value, followed closely by HP-FS-PSO (differing only in the third decimal place). These performances are again superior to M1, supporting the idea of MLP hyperparameter tuning and feature selection influence.

However, differently from Table~\ref{tab:summary}, HP-FS-PSO was not statistically better than M3, as observed in the $p$-value $>0.05$. Therefore, the improvement obtained during the validation steps was not enough to provide a statistical difference, considering the test data. The high standard deviation over the testing set explains such behavior, which was considerably smaller regarding the optimization steps. 

\section{Conclusion}
\label{s.conclusions}

This paper analyzed the compound problem of ANN hyperparameters tuning and feature selection for the quality classification of wood boards in the sawmill industry. Experiments showed that a solution based on PSO led to satisfactory results compared to baseline methods. According to a statistical test, results show a significant difference during the optimization task but not for the generalization phase. 


These experimental results suggest that MLP hyperparameter tuning and feature selection are essential to obtain models with higher predictive performance. Also, one can notice that these tasks are interdependent since the hyperparameter values should be adjusted according to a subset of features and vice versa. Consequently, for the problem investigated in this work, performing only one of them was enough to reach substantial gain. Finally, the accuracy obtained in this study supports employing machine learning models for industrial implementation, contributing to overall cost reduction and improvement in competitiveness. Regarding future works, we intend to perform a transfer learning from a CNN trained using a dataset composed of a more substantial number of wood image samples. Besides, we are willing to investigate and compare different image descriptors, non-linear data reduction techniques, and deep learning models.
%
%
%
\bibliographystyle{splncs04}
\bibliography{references}

\begin{thebibliography}{10}
\providecommand{\url}[1]{\texttt{#1}}
\providecommand{\urlprefix}{URL }
\providecommand{\doi}[1]{https://doi.org/#1}

\bibitem{abdullah2007}
Abdullah, A., Ismail, N.K.N., Kadir, T.A.A., Zain, J.M., Jusoh, N.A., Ali,
  N.M.: Agar wood grade determination system using image processing technique.
  Proceedings of the International Conference on Electrical Engineering and
  Informatics Institut Teknologi Bandung  (2007)

\bibitem{affonso2017}
Affonso, C., Rossi, A.L.D., Vieira, F.H.A., Carvalho, A.C.P.L.F.: Deep learning
  for biological image classification. Expert Systems With Applications
  \textbf{85},  114--122 (2017).
  \doi{http://dx.doi.org/10.1016/j.eswa.2017.05.039}

\bibitem{brodersen2010}
Brodersen, K.H., Ong, C.S., Stephan, K.E., Buhmann, J.M.: The balanced accuracy
  and its posterior distribution. In: 2010 20th International Conference on
  Pattern Recognition. pp. 3121--3124 (Aug 2010). \doi{10.1109/ICPR.2010.764}

\bibitem{cao2021new}
Cao, Y., Zandi, Y., Rahimi, A., Wu, Y., Fu, L., Wang, Q., Deni{\'c}, N.,
  Khadimallah, M.A., Mili{\v{c}}, M., Paunovi{\'c}, M.: A new intelligence
  fuzzy-based hybrid metaheuristic algorithm for analyzing the application of
  tea waste in concrete as natural fiber. Computers and Electronics in
  Agriculture  \textbf{190},  106420 (2021)

\bibitem{de2021fine}
De~Souza, L.A., Passos, L.A., Mendel, R., Ebigbo, A., Probst, A., Messmann, H.,
  Palm, C., Papa, J.P.: Fine-tuning generative adversarial networks using
  metaheuristics-a case study on barrett's esophagus identification. In:
  Bildverarbeitung f{\"u}r die Medizin. pp. 205--210 (2021)

\bibitem{gu2009}
Gu, I.Y.H., Andersson, H., Vicen, R.: Automatic classification of wood defects
  using support vector machines. In: Computer Vision and Graphics. pp.
  356--367. Springer Berlin Heidelberg (2009)

\bibitem{hall2009}
Hall, M., Frank, E., Holmes, G., Pfahringer, B., Reutemann, P., Witten, I.H.:
  The weka data mining software: An update. SIGKDD Explor. Newsl.
  \textbf{11}(1),  10--18 (2009). \doi{10.1145/1656274.1656278}

\bibitem{haralick1973}
Haralick, R., Shanmugam, K., Distein, I.: Textual features for image
  classification. IEEE Transaction on Systems, Man and Cybernetics
  \textbf{SMC-3}(6),  610--621 (1973)

\bibitem{kennedy1997}
Kennedy, J., Eberhart, R.C.: A discrete binary version of the particle swarm
  algorithm. In: 1997 IEEE International Conference on Systems, Man, and
  Cybernetics. Computational Cybernetics and Simulation. vol.~5, pp. 4104--4108
  (Oct 1997). \doi{10.1109/ICSMC.1997.637339}

\bibitem{kennedy2001}
Kennedy, J., Eberhart, R.: Swarm Intelligence. Morgan Kaufmann Publishers
  (2001)

\bibitem{kennedy1995}
Kennedy, J., Eberhart, R.: Particle swarm optimization. In: Proceedings of the
  IEEE International Conference on Neural Networks. vol.~4, pp. 1942--1948.
  Perth, Australia (1995)

\bibitem{luo2016}
Luo, G.: A review of automatic selection methods for machine learning
  algorithms and hyper-parameter values. Network Modeling Analysis in Health
  Informatics and Bioinformatics  \textbf{5}(1), ~18 (May 2016).
  \doi{0.1007/s13721-016-0125-6}

\bibitem{ojala1996}
Ojala, T., Pietikainen, M., Harwood, D.: Comparative study of texture measures
  with classification based on feature distributions. Pattern Recognit pp.
  51--59 (1996)

\bibitem{passosSIBGRAPI:2017}
Passos, L.A., ao~Paulo~Papa, J.: Fine-tuning infinity restricted boltzmann
  machines. In: 2017 30th SIBGRAPI Conference on Graphics, Patterns and Images
  (SIBGRAPI). pp. 63--70. IEEE (2017)

\bibitem{pereiraECCOMAS:19}
Pereira, C.R., Passos, L.A., Rodrigues, D., de~Souza, A.N., Papa, J.P.:
  Jade-based feature selection for non-technical losses detection. In: ECCOMAS
  Thematic Conference on Computational Vision and Medical Image Processing. pp.
  141--156. Springer (2019)

\bibitem{pham1996}
Pham, D.T., Alcock, R.J.: Automatic detection of defects on birch wood boards.
  Proceedings of the Institution of Mechanical Engineers, Part E: Journal of
  Process Mechanical Engineering  \textbf{210}(1),  45--52 (1996).
  \doi{10.1243/0954408991529852}

\bibitem{qi2018}
Qi, C., Fourie, A., Chen, Q.: Neural network and particle swarm optimization
  for predicting the unconfined compressive strength of cemented paste
  backfill. Construction and Building Materials  \textbf{159},  473 -- 478
  (2018). \doi{https://doi.org/10.1016/j.conbuildmat.2017.11.006}

\bibitem{r2014}
{R Core Team}: R: A Language and Environment for Statistical Computing. R
  Foundation for Statistical Computing, Vienna, Austria (2014)

\bibitem{roder2021reinforcing}
Roder, M., Passos, L.A., de~Rosa, G.H., de~Albuquerque, V.H.C., Papa, J.P.:
  Reinforcing learning in deep belief networks through nature-inspired
  optimization. Applied Soft Computing  \textbf{108},  107466 (2021)

\bibitem{roder2020harnessing}
Roder, M., de~Rosa, G.H., Passos, L.A., Papa, J.P., Rossi, A.L.D.: Harnessing
  particle swarm optimization through relativistic velocity. In: 2020 IEEE
  Congress on Evolutionary Computation (CEC). pp.~1--8. IEEE (2020)

\bibitem{roder2017}
Roder, M., Rossi, A.L.D., de~Oliveira~Affonso, C.: Boosting machine learning
  techniques for wood quality classification by particle swarm optimization.
  In: Encontro Nacional de Intelig\^encia Artificial e Computacional. Sociedade
  Brasileira de Computação (2017)

\bibitem{rodrigues2020adaptive}
Rodrigues, D., de~Rosa, G.H., Passos, L.A., Papa, J.P.: Adaptive improved
  flower pollination algorithm for global optimization. In: Nature-Inspired
  Computation in Data Mining and Machine Learning, pp. 1--21. Springer (2020)

\bibitem{shi1998}
Shi, Y., Eberhart, R.: A modified particle swarm optimizer. In: 1998 IEEE
  International Conference on Evolutionary Computation Proceedings. IEEE World
  Congress on Computational Intelligence (Cat. No.98TH8360). pp. 69--73 (May
  1998). \doi{10.1109/ICEC.1998.699146}

\bibitem{tiryaki2014}
Tiryaki, S., Malkoçoğlu, A., Şükrü Özşahin: Using artificial neural
  networks for modeling surface roughness of wood in machining process.
  Construction and Building Materials  \textbf{66},  329 -- 335 (2014).
  \doi{https://doi.org/10.1016/j.conbuildmat.2014.05.098}

\bibitem{vieira2016}
Vieira, F.H.A.: Image processing through machine learning for wood quality
  classification. Ph.D. thesis, Faculdade de Engenharia de Guaratinguetá
  (FEG), UNESP (2016)

\bibitem{Wilcoxon:45}
Wilcoxon, F.: Individual comparisons by ranking methods. Biometrics Bulletin
  \textbf{1}(6),  80--83 (1945)

\end{thebibliography}
%


\end{document}